\documentclass{article}
\usepackage{times}
\usepackage{authblk}

\usepackage[bottom=1.in,top=0.8in, ]{geometry}

\usepackage{amsmath}
\usepackage{amsfonts}
\usepackage{graphicx}
\usepackage{tcolorbox}
\usepackage{hyperref}
\usepackage{tabu}

\usepackage{layouts}

\usepackage{booktabs}
\usepackage{tikz}
\usepackage{subcaption}

\usepackage{amsmath}
\usepackage{graphicx}
\usepackage{float}
\usepackage{cleveref}
\crefformat{equation}{(#2#1#3)}
\crefrangeformat{equation}{(#3#1#4) to~(#5#2#6)}
\crefmultiformat{equation}{(#2#1#3)}%
{ and~(#2#1#3)}{, (#2#1#3)}{ and~(#2#1#3)}

\newcommand{\R}{\mathbb{R}}

\newcommand{\N}{\mathbb{N}}
\newcommand{\C}{\mathbb{C}}

\newcommand{\Id}{\operatorname{Id}}

\newcommand{\calP}{\mathcal{P}}

\newcommand{\calV}{\mathcal{V}}

\newcommand{\norm}[1]{\left\Vert #1 \right\Vert}
\newcommand{\set}[1]{\left\lbrace #1\right\rbrace}

\DeclareMathOperator*{\argmax}{arg\,max}
\DeclareMathOperator*{\argmin}{arg\,min}

\usepackage{mathtools, dirtytalk}

\newcounter{RonCounter}

\newcounter{StefanCounter}

\newcounter{DACounter}

\newcounter{JoanCounter}

\begin{document}

\newtheorem{definition}{Definition}
\newtheorem{example}{Example}

%\title{A Rate-Distortion Framework for Explaining Black-box Model Decisions \thanks{Supported by organization x.}}
\title{A Rate-Distortion Framework for Explaining Black-box Model Decisions}
\date{}
%\author{First Author\inst{1}\orcidID{0000-1111-2222-3333} \and
%Second Author\inst{2,3}\orcidID{1111-2222-3333-4444} \and
%Third Author\inst{3}\orcidID{2222--3333-4444-5555}}
%

\author[1]{Stefan Kolek}
\author[1]{Duc Anh Nguyen}
\author[1]{Ron Levie}
\author[2]{Joan Bruna}
\author[1]{Gitta Kutyniok}
\affil[1]{Department of Mathematics, Ludwig Maximilian University, Munich}
\affil[2]{Courant Institute of Mathematical Sciences, New York University, New York}

\maketitle
\begin{abstract}
We present the \emph{Rate-Distortion Explanation} (RDE) framework, a mathematically well-founded method for explaining black-box model decisions. The framework is based on perturbations of the target input signal and applies to any differentiable pre-trained model such as neural networks. Our experiments demonstrate the framework's adaptability to diverse data modalities, particularly images, audio, and physical simulations of urban environments.

%\keywords{First keyword  \and Second keyword \and Another keyword.}
\end{abstract}
%
%
%
%\DA{orcidID in authors? don't know what it is. IN general check the format of authors}

\section{Introduction} % Inutuitive intro to RDE and comparison to related methods 
Powerful machine learning models such as deep neural networks are inherently opaque, which has motivated numerous explanation methods  that the research community developed over the last decade \cite{Layerwise_relevance_prop2015,Smooth_Grad_2017,Integrated_gradient_2017_sundararajan,LIME_2016,SHAP_neurips_2017,RDE_original_2019,Fong_vedaldi_2017,chang_GAN_explanation_2018}. The meaning and validity of an explanation depends on the underlying principle of the explanation framework. Therefore, a trustworthy explanation framework must align intuition with mathematical rigor while maintaining maximal flexibility and applicability. We believe the \emph{Rate-Distortion Explanation} (RDE) framework, first proposed by \cite{RDE_original_2019}, then extended by \cite{in_distribution_cosmas_2020}, as well as the similar framework in \cite{chang_GAN_explanation_2018}, meets the desired qualities. In this chapter, we aim to present the RDE framework in a revised and holistic manner. Our generalized RDE framework can be applied to any model (not just classification tasks), supports in-distribution interpretability (by leveraging in-painting GANs), and admits interpretation queries (by considering suitable input signal representations). 
 
 The typical setting of a (local) explanation method is given by a pre-trained model $\Phi:\R^n\to\R^m,$
  and a data instance $x\in\R^n$. The model $\Phi$ can be either a classification task with $m$ class labels or a regression task with $m$-dimensional model output. The model decision $\Phi(x)$ is to be explained. In the original RDE framework \cite{RDE_original_2019}, an explanation for $\Phi(x)$ is a set of feature components $S\subset \set{1,\hdots,n}$ in $x$ that are deemed relevant for the decision $\Phi(x)$. The core principle behind the RDE framework is that a set $S\subset \set{1,\hdots,n}$ contains all the relevant components if $\Phi(x)$ remains (approximately) unchanged after modifying $x_{S^c}$, i.e., the components in $x$ that are not deemed relevant. In other words, $S$ contains all relevant features if they are sufficient for producing the output $\Phi(x)$. To convey concise explanatory information, one aims to find the \emph{minimal set} $S\subset \set{1,\hdots,n}$ with all the relevant components.
 As demonstrated in \cite{RDE_original_2019} and \cite{complexity_ana_2019}, the minimal relevant set $S\subset \set{1,\hdots,n}$ cannot be found combinatorically in an efficient manner for large input sizes. A meaningful approximation can nevertheless be found by optimizing a sparse continuous mask $s\in [0,1]^n$ that has no significant effect on the output $\Phi(x)$ in the sense  that 
 $\Phi(x)\approx \Phi(x\odot s + (1-s)\odot v)$
 should hold for appropriate perturbations $v\in\R^n$, where $\odot$ denotes the componentwise multiplication. Suppose  $d\big(\Phi(x),\Phi(y)\big)$ is a measure of distortion (e.g. the $\ell_2$-norm) between the model outputs for $x,y\in\R^n$ and $\mathcal{V}$  is a distribution over appropriate perturbations $v\sim \mathcal{V}$. An explanation in the RDE framework can be found as a solution mask $s^*$ to the following minimization problem:
 \begin{align*}
     s^* \coloneqq \quad \argmin_{s\in[0,1]^n} \mathop{\mathbb{E}}_{v\sim \mathcal{V}}\Bigg[d\Big(\Phi(x),\Phi(x\odot s + (1-s)\odot v)\Big)\Bigg] + \lambda \|s\|_1,
 \end{align*}
 where $\lambda>0$ is a hyperparameter controlling the sparsity of the mask. 

%\joan{Perhaps using a linear decomposition for $x$ and precising that $S$ indexes a subset of basis coordinates is more clear. As is, it is not entirely clear what is meant by $S$.} \stefan{I don't understand how we can use a $x_S$ notation here since in the formula above we relax S to continuous masks $s\in[0,1]^n$. Can someone help me understand?}
%\joan{I would suggest a figure here that illustrates the principle of RDE geometrically. We want to identify a \emph{stable} subspace of $\Phi$ at $x$. The salient features retained in $S$ encode this subspace (as normal directions). Relate to gradient-based methods (which also admit a similar geometric interpretation)}\stefan{I like the idea of a figure however we are limited to an even number of pages and another figure would require a good amount of additional editing. Since the figure and the stable subspace view are particularly interesting in the context of robustness, I would suggest to add a figure when we write down the next experiments in the next paper.}
 
 We further generalize the RDE framework to abstract input signal representations $x=f(h)$, where $f$ is a data representation function with input $h$. The philosophy of the generalized RDE framework is that an explanation for generic input signals $x=f(h)$ should be some simplified version of the signal, which is interpretable to humans. This is achieved by demanding sparsity in a suitable representation system $h$, which ideally optimally represents the class of explanations that are desirable for the underlying domain and interpretation query. This philosophy underpins our experiments on image classification in the wavelet domain, on audio signal classification in the Fourier domain, and on radio map estimation in an urban environment domain. Therein we demonstrate the versatility of our generalized RDE framework.

\section{Related works}

To our knowledge, the explanation principle of optimizing a mask $s\in [0,1]^n$ has been first proposed in \cite{Fong_vedaldi_2017}. Fong et al. \cite{Fong_vedaldi_2017} explained  image classification decisions by considering one of the two \say{deletion games}: (1) optimizing for the smallest deletion mask that causes the class score to drop significantly or (2) optimizing for the largest deletion mask that has no significant effect on the class score.  The original RDE approach \cite{RDE_original_2019} is based on the second deletion game and connects the deletion principle to rate-distortion-theory, which studies lossy data compression. Deleted entries in \cite{Fong_vedaldi_2017}  were replaced with either constants, noise, or blurring and deleted entries in \cite{RDE_original_2019} were replaced with noise.

Explanation methods introduced before the \say{deletion games} principle from \cite{Fong_vedaldi_2017}  were typically based upon gradient-based methods \cite{Smooth_Grad_2017}\cite{Integrated_gradient_2017_sundararajan}, propagation of activations in neurons \cite{Layerwise_relevance_prop2015}\cite{DeepLIFT_2017}, surrogate models \cite{LIME_2016}, and game-theory \cite{SHAP_neurips_2017}. Gradient-based methods such as smoothgrad \cite{Smooth_Grad_2017} suffer from a lacking principle of relevance beyond local sensitivity. Reference-based methods such as Integrated Gradients \cite{Integrated_gradient_2017_sundararajan} and DeepLIFT \cite{DeepLIFT_2017} depend on a reference value, which has no clear optimal choice. DeepLIFT and LRP assign relevance by propagating neuron activations, which makes them dependent on the implementation of $\Phi$. LIME \cite{LIME_2016} uses an interpretable surrogate model that approximates $\Phi$ in a neighborhood around $x$.  Surrogate model explanations are  inherently limited for complex models $\Phi$ (such as image classifiers) as they  only admit very local approximations. Generally, explanations that only depend on the model behavior on a small neighborhood $U_x$ of $x$ offer limited insight. Lastly, Shapley values-based explanations \cite{SHAP_neurips_2017} are grounded in Shapley values from game-theory. They assign relevance scores as weighted averages of marginal contributions of respective features. Though Shapley values are mathematically well-founded, relevance scores cannot be computed exactly for common input sizes such as $n\geq 50$, since one exact relevance score generally requires $O(2^n)$ evaluations of $\Phi$ \cite{fast-hierarchichal-games}. 

A notable difference between the RDE method and additive feature explanations  \cite{SHAP_neurips_2017} is that the values in the mask $s^*$ do not add up to the model output. The additive property as in \cite{SHAP_neurips_2017} takes the view that features individually contribute to the model output and relevance should be reflected by their contributions. We emphasize that the RDE method is designed to look for a \emph{set} of relevant features and \emph{not} an estimate of individual relative contributions. This is particularly desirable when only groups of features are interpretable, as for example in image classification tasks, where individual pixels do not carry any interpretable meaning.
Similarly to Shapley values, the explanation in the RDE framework cannot be computed exactly, as it requires solving a non-convex minimization problem. However, the RDE method can take full advantage of modern optimization techniques. Furthermore, the RDE method is a model-agnostic explanation technique, with a mathematically principled and intuitive notion of relevance as well as enough flexibility to incorporate the model behavior on meaningful input regions of $\Phi$.

The meaning of an explanation based on deletion masks $s\in [0,1]^n$ depends on the nature of the perturbations that replace the deleted regions. Random \cite{RDE_original_2019} \cite{Fong_vedaldi_2017} or blurred \cite{Fong_vedaldi_2017} replacements  $v\in \R^n$ may result in a data point $x\odot s + (1-s)\odot v$ that falls out of the natural data manifold on which $\Phi$ was trained on. This is a subtle though important problem, since such an explanation may depend on evaluations of $\Phi$ on data points from undeveloped decision regions. The latter motivates \emph{in-distribution interpretability}, which considers meaningful perturbations that keep $x\odot s + (1-s)\odot v$ in the data manifold. \cite{chang_GAN_explanation_2018} was the first work that suggested to use an inpainting-GAN to generate meaningful perturbations to the \say{deletion games}. The authors of \cite{in_distribution_cosmas_2020} then applied in-distribution interpretability to the RDE method in the challenging modalities  \emph{music} and \emph{physical simulations of urban environments}. Moreover, they demonstrated that the RDE method in \cite{RDE_original_2019} can be extended to answer so-called \say{\emph{interpretation queries}}. For example, the RDE method was applied in \cite{in_distribution_cosmas_2020} to an instrument classifier to answer the global interpretation query \say{\emph{Is magnitude or phase in the signal more important for the classifier?}}. Most recently, in \cite{kolek2021cartoon}, we introduced CartoonX as a novel explanation method for image classifiers, answering the interpretation query \emph{``What is the relevant piece-wise smooth part of an image?"} by applying RDE in the wavelet basis of images.

\section{Rate-distortion explanation framework}
%Furthermore, we define the \emph{rate-distortion function} $R \colon \R_+ \to [d]$ by
%\begin{align*}
  %  R(\epsilon) = \min \set{\abs{S}: S \subseteq [d], \;D(S) \leq \epsilon}, \;\epsilon \geq 0.
%\end{align*}
Based on the original RDE approach from \cite{RDE_original_2019}, in this section, we present a general formulation of the RDE framework and discuss several implementations. While \cite{RDE_original_2019} focuses merely on image classification with explanations in pixel representation, we will apply the RDE framework not only to more challenging domains but also to different input signal representations. Not surprisingly, the combinatorical optimization problem in the RDE framework, even in simpler form, is extremely hard to solve \cite{RDE_original_2019} \cite{complexity_ana_2019}. This motivates heuristic solution strategies, which will be discussed in Subsection \ref{section:implementation}.

\subsection{General formulation}\label{sec:general_formulation}
%We begin with the definition of \emph{obfuscations} and \emph{expected distortion}, which generalize the notion of obfuscation and distortion from the original RDE formulation in \cite{RDE_original_2019}.
%\ron{I think that before you go into the definitions it is important to motivate the notion of  data representation. Perhaps give and example, e.g., pixel representation, basis representation, and decoder.}

It is well-known that in practice there are different ways to describe a signal $x \in \R^n$. Generally speaking, $x$ can be represented by a data representation function $f:\prod_{i=1}^k\R^{d_i}\to\R^n$,
\begin{align} \label{eq:abstract_representation}
    x = f(h_1, \hdots, h_k),
\end{align}
for some inputs $h_i \in \R^{d_i}$, $d_i \in \N$, $i \in \set{1, \hdots, k}$, $k\in \N$. Note, we do not restrict ourselves to linear data representation functions $f$. To briefly illustrate the generality of this abstract representation, we consider the following examples.

%\paragraph{Data representation and interpretation queries.}
%We reformulated the RDE framework for general data representations $x=f(h_1,...,h_k)$ for maximal flexibility with respect to interpretation queries. First we would like to briefly illustrate how $x=f(h_1,...,h_k)$ is indeed a generalized data representation: \ron{Sentences never end with colon. Just finish with a period.}
\begin{example}[Pixel representation]\label{Example:pixel_representation}
    %\stefan{is \say{euclidean representation} accurate or should i use a differnet title?} \ron{pixel representation? Pixel space representation?} \DA{I think pixel representation is only used for images? How about entrywise or componentwise representation?} \ron{True, but most of the most well known applictations are in images. I think it should be clear, and, also, this is just an example. Another option is standard basis representation.} 
    An arbitrary (vectorized) image $x \in \R^n$ can be simply represented pixelwise 
    \begin{align*}
        x = \begin{bmatrix} x_1 \\ \vdots \\ x_n\end{bmatrix} = f(h_1, \hdots, h_n),
    \end{align*}
    with $h_i \coloneqq x_i$ being the individual pixel values and $f \colon \R^n \to \R^n$ being the identity transform. 
    
    %\ron{I don't think that here $f$ is the image. $f$ is the identity transform, and $h_1\in \mathbb{R}^d$ is the image (or use $h_1,\ldots,h_d\in\mathbb{R}$ as the pixel values).}
\end{example}
Due to its simplicity, this standard basis representation is a reasonable choice when explaining image classification models. However, in many other applications, one requires more sophisticated representations of the signals, such as through a possibly redundant dictionary. 
\begin{example}%[Discrete Fourier Transform]
\label{ex:FFT}
    Let $\set{\psi_j}_{j =1}^k$, $k \in \N$, be a dictionary in $\R^n$, e.g., a basis. A signal $x \in \R^n$ is represented as
    \begin{align*}
        x = \sum_{j =1}^k h_j \psi_j,
    \end{align*}
    where $h_j\in \R$, $j \in \set{1, \hdots, k}$, are appropriate coefficients. In terms of the abstract representation \cref{eq:abstract_representation}, we have $d_j = 1$ for $j \in \set{1, \hdots, k}$ and $f$ is the function that yields the weighted sum over $\psi_j$. Note that \Cref{Example:pixel_representation} can be seen as a special case of this representation. 
\end{example}
The following gives an example of a non-linear representation function $f$.
\begin{example}
\label{ex:FFT1}
    Consider the discrete inverse Fourier transform, defined as %\DA{changed the image set to $\C^n$ instead of $\R^n$}
    \begin{align*}
        &f: \prod_{j=1}^{n}\R_+ \times \prod_{j=1}^n[0,2\pi]  \to \mathbb{C}^n,\\ &\big[f(m_1,...,m_n,\omega_1,...,\omega_n)\big]_l\coloneqq \frac{1}{n} \sum_{j=1}^{n} \underbrace{m_je^{i\omega_j}}_{\coloneqq c_j\in\C}e^{i2\pi  l(j-1)/n}, \; l \in \set{1, \hdots, n},
    \end{align*}
    where $m_j$ and $\omega_j$ are respectively the magnitude and the phase of the $j$-th discrete Fourier coefficient $c_j$. Thus every signal $x \in \R^n \subseteq \C^n$ can be represented in terms of \cref{eq:abstract_representation} with $f$ being the discrete inverse Fourier transform while $h_{j}$, $j=1,\hdots,k$ (with $k=2n$) being specified as $m_{j'}$ and $\omega_{j'}$, $j' = 1, \hdots, n$. 
    %where  $c_j$, $j \in \set{1, \hdots, n}$ is the Fourier coefficient that encodes the magnitude $m_k$ and the phase $\omega_k$ of the sinusoidal component $e^{i2\pi  na/k}$ in the signal $f(m_1,...,m_k,\omega_1,...,\omega_k)_n$.
\end{example}
Further examples of dictionaries $\set{\psi_j}_{j=1}^k$ include the discrete wavelet \cite{Romberg06wavelet-domainapproximation}, cosine \cite{cosine-transform} or shearlet \cite{kutyniok2010} representation systems and many more. In these cases, the coefficients $h_i$ are given by the forward transform and $f$ is referred to as the backward transform. 
%Note how the previous example of the discrete Fourier transform used the dimensions $d_i=1$. Representations $x=f(h_1,...,h_k)$ with $h_i\in\R^{d_i}$ and $d_i>1$ become particularly interesting for so-called \emph{interpretation queries}. We illustrate this with an example.
Note that in the above examples we have $d_i = 1$, i.e., the input vectors $h_i$ are real-valued. In many situations, one is also interested in representations $x = f(h_1, \hdots, h_k)$ with $h_i \in \R^{d_i}$ where $d_i >1$. 
\begin{example}%[Interpretation query: is magnitude or phase more important?]
\label{ex:query_phase_magnitude}
    Let $k=2$ and define $f$ again as the discrete inverse Fourier transform, but as a function of two components: (1) the entire magnitude spectrum and (2) the entire frequency spectrum, namely
    \begin{align*}
        &f: \R_+^n \times [0,2\pi]^n,\\
        &\big[f(m,\omega)\big]_l \coloneqq \frac{1}{n} \sum_{j=1}^{n} \underbrace{m_j e^{i\omega_j}}_{\coloneqq c_n\in\C}e^{i2\pi  l(j-1)/n},\; l \in \set{1, \hdots, n}.
    \end{align*}
    Similarly, instead of individual pixel values, one can consider patches of pixels in an image $x \in \R^n$ from \Cref{Example:pixel_representation} as the input vectors $h_i$ to the identity transform $f$. We will come back to these examples in the experiments in \Cref{sec:experiments}.
    %For such a representation, a mask $s\in [0,1]^k = [0,1]^2$ turns off or on the entire magnitude or phase spectra. We can then explore whether the magnitude or phase tends to be more important for a model $\Phi: \R^d\to \R^m$ by minimizing the expected distortion over
    %all data $x\sim\mathcal{D}$:
    %$$
    %s^*(\lambda) \coloneqq \argmin_{s\in[0,1]^2} \mathop{\mathbb{E}}_{x\sim\mathcal{D}} D(x,s,\mathcal{V}_s, \Phi) 
    %$$
    %\ron{since $s$ here is only 2D, the formulation with l1 regularization does not make sense. You need only compare two versions of he mask: masking magnitude, and masking phase.}
    %\stefan{In the in-distribution paper this was the approach to: is magnitude or phase more important. I find it a bit unrealistic to turn off the entire spectrum. Why not determine if phase or magnitdue is important by computing several heatmaps for data points $x\sim\mathcal{D}$ and seeing if more magnitude or more phase components are important?} \ron{What you suggest is also a valid approach, but we already have the old implementation, so let's keep it.}
\end{example}
Finally, we would like to remark that our abstract representation
$$x = f(h_1,\hdots,h_k)$$
also covers the cases where the signal is the output of a decoder or generative model $f$ with inputs $h_1, \hdots, h_k$ as the code or the latent variables. 

As was discussed in previous sections, the main idea of the RDE framework is to extract the relevant features of the signal based on the optimization over its perturbations defined through masks. The ingredients of this idea are formally defined below.
\begin{definition}[Obfuscations and expected distortion]\label{def: distortion and obsfustation}
Let $\Phi:\R^n\to\R^m$ be a model and $x\in \R^n $ a data point with a data representation $x =f(h_1,...,h_k)$ as discussed above. For every mask $s\in[0,1]^k$, let $\calV_s$ be a probability distribution over $\prod_{i=1}^k\R^{d_i}$. Then the \emph{obfuscation} of $x$ with respect to $s$ and $\mathcal{V}_s$ is defined as the random vector
$$
y \coloneqq f(s\odot h + (1-s)\odot v),
$$
where $v\sim\mathcal{V}_s$, $(s\odot h)_i = s_i h_i\in\R^{d_i}$ and $((1-s)\odot v)_i= (1-s_i)v_i\in\R^{d_i}$ for $i\in\set{1, \hdots, k}$. Furthermore, the \emph{expected distortion} of $x$ with respect to the mask $s$ and the perturbation distribution $\mathcal{V}_s$ is defined as
$$
D(x,s,\mathcal{V}_s, \Phi)\coloneqq \mathop{\mathbb{E}}_{v\sim \mathcal{V}_s} \Bigg[ d\Big(\Phi(x), \Phi(y)\Big)\Bigg],
$$
where $d:\R^m\times \R^m\to\R_+$ is a measure of distortion between two model outputs.
\end{definition}
In the RDE framework, the explanation is given by a mask that minimizes distortion while remaining relatively sparse. The rate-distortion-explanation mask is defined in the following.
\begin{definition}[The RDE mask]\label{def:rde mask}
In the setting of Definition \ref{def: distortion and obsfustation} we define the \emph{RDE mask} as a solution $s^*(\ell)$ to the minimization problem
\begin{align}\label{eq:rde mask}
    \min_{s\in \{0,1\}^k} \quad D(x,s,\mathcal{V}_s, \Phi) \quad \text{ s.t. } \quad \norm{s}_0 \leq \ell,
\end{align}
where $\ell\in \set{1, \hdots, k}$ is the desired level of sparsity.
\end{definition}
Here, the RDE mask is defined as the binary mask that minimizes the expected distortion while keeping the sparsity smaller than a certain threshold. Besides this, one could obviously also define the RDE mask as the sparsest binary mask that keeps the distortion lower than a given threshold, as defined in \cite{RDE_original_2019}. 
Geometrically, one can interpret the RDE mask as a subspace that is stable under $\Phi$. If $x=f(h)$ is the input signal and $s$ is the RDE mask for $\Phi(x)$ on the coefficients $h$, then the associated subspace $R_\Phi(s)$ is defined as the space of feasible obfuscations of $x$ with $s$ under $\calV_s$, i.e.,
$$
R_\Phi(s) \coloneqq\{f(s\odot h + (1-s)\odot v)\;|\;v\in\text{supp}\calV_s \},
$$
where $\text{supp}\calV_s$ denotes the support of the distribution $\calV_s$. The model $\Phi$ will act similarly on signals in $R_\Phi(s)$ due to the low expected distortion $ D(x,s,\mathcal{V}_s, \Phi)$---making the subspace stable under $\Phi$. Note that RDE directly optimizes towards a subspace that is stable under $\Phi$. If, instead, one would choose the mask $s$ based on information of the gradient $\nabla\Phi(x)$ and Hessian $\nabla^2\Phi(x)$, then only a local neighborhood around $x$ would tend to be stable under $\Phi$ due to the local nature of the gradient and Hessian.
Before discussing practical algorithms to approximate the RDE mask in Subsection \ref{section:implementation}, we will review frequently used obfuscation strategies, i.e., the distribution $\calV_s$, and measures of distortion. 

%\joan{Again, it would be helpful to complement with a geometric description of the mask in terms of a stable subspace. I am pretty sure this can be interpreted in terms of robust derivatives. In other words, how are the non-local directions found by the RDE mask different from the first $\|s\|_0$ eigenvectors of the Hessian $\nabla^2 \Phi(x)$? This needs to be explained at the intuitive level at least. This will motivate better the next section. }
%\stefan{I added a couple of lines above to convey the intuition of RDE as a stable subspace.}

\subsubsection{Obfuscation strategies and in-distribution interpretability.} 
The meaning of an explanation in RDE depends greatly on the nature of the perturbations $v\sim\mathcal{V}_s$. A particular choice of $\mathcal{V}_s$ defines an \emph{obfuscation strategy}. Obfuscations are either \emph{in-distribution}, i.e., if the obfuscation
$$
f(s\odot h + (1-s)\odot v)
$$
lies on the natural data manifold that $\Phi$ was trained on, or \emph{out-of-distribution} otherwise. Out-of-distribution obfuscations pose the following problem. The RDE mask (see Definition \ref{def:rde mask}) depends on evaluations of $\Phi$ on obfuscations $f(s\odot h + (1-s)\odot v)$. If $f(s\odot h + (1-s)\odot v)$ is not on the natural data manifold that $\Phi$ was trained on, then it may lie in undeveloped regions of $\Phi$. In practice, we are interested in explaining the behavior of $\Phi$ on realistic data and an explanation can be corrupted if $\Phi$ did not develop the region of out-of distribution points $f(s\odot h + (1-s)\odot v)$. One can guard against this by choosing $\mathcal{V}_s$ so that $f(s\odot h + (1-s)\odot v)$ is in-distribution.  %However, this is non-trivial.
Choosing $\mathcal{V}_s$ in-distribution boils down to modeling the conditional data distribution -- a non-trivial task.

\begin{example}[In-distribution obfuscation strategy]\label{ex:GAN_obfuscation}
    In light of the recent success of generative adversarial networks (GANs) in generative modeling \cite{goodfellow2014}, one can train an in-painting GAN \cite{Generative_inpainting_2017}
    $$
    G(h,s,z)\in \prod_{i=1}^k \R^{d_i},
    $$
    where $z$ are random latent variables of the GAN, such that the obfuscation $f\big(s\odot h + (1-s)\odot G(h,s,z) \big)$ lies on the natural data manifold (see also \cite{chang_GAN_explanation_2018}). In other words, one can choose $\calV_s$ as the distribution of $v\coloneqq G(h,s,z)$, where the randomness comes from the random latent variables $z$.
\end{example} 

\begin{example}[Out-of-distribution obfuscation strategies]
    A very simple obfuscation strategy is Gaussian noise. In that case, one defines $\mathcal{V}_s$ for every $s\in[0,1]^k$ as
    $$
    \mathcal{V}_s\coloneqq \mathcal{N}(\mu,\Sigma), 
    $$
    where $\mu$ and $\Sigma$ denote a pre-defined mean vector and covariance matrix. In \Cref{section:experiment_images}, we give an example of a reasonable choice for $\mu$ and $\Sigma$ for image data.
    Alternatively, for images with pixel representation (see \Cref{Example:pixel_representation}) one can mask out the deleted pixels by blurred inputs, $v = K*x$, where $K$ is a suitable blur kernel.
    %\stefan{I wanted to keep it sure but I am not sure if I wrote down the blurring operation correctly with an abstract kernel. Let me know if not.} \ron{Perhaps write it as convolution $h*K$ or as $v_j \coloneqq \sum_{i=1}^k h_i K_{i-j}$?}
\end{example}
\begin{table}[h!]
    \begin{center}
    \begin{tabular}{ |c|c|c| } 
    \hline
     {\bf Obfuscation strategy} & {\bf Perturbation formula} & {\bf In-distribution}\\
     \hline
     Constant & $v\in\R^d$ & -- \\ 
     \hline
     Noise &    $v\sim\mathcal{N}(\mu,\Sigma)$ & --\\ 
     \hline
     Blurring &  $v= K*x$ & --\\ 
     \hline
     Inpainting-GAN &  $v = G(h,s,z)$& \checkmark
     \\
     \hline
    \end{tabular}
    \vspace{0.3cm}
    \caption{Common obfuscation strategies with their perturbation formulas.}
    \label{tab:obfuscation strategies}
    \end{center}
\end{table}
We summarize common obfuscation strategies for a given target signal in \Cref{tab:obfuscation strategies}.

\subsubsection{Measure of distortion.}
 Various options exist for the measure $d \colon \R^m \times \R^m \to \R$ of the distortion between model outputs. The measure of distortion should be chosen according to the task of the model $\Phi:\R^n\to\R^m$ and the objective of the explanation.

\begin{example}[Measure of distortion for classification task]\label{example:classification_measure}
    Consider a classification model $\Phi:\R^n\to\R^m$ and a target input signal $x \in \R^n$. The model $\Phi$ assigns to each class $j\in\set{1, \hdots, m}$ a (pre-softmax) score $\Phi_j(x)$ and the predicted label is given by $j^*\coloneqq \argmax_{j \in \set{1, \hdots, m}} \Phi_j(x)$. 
    One commonly used measure of the distortion between the outputs at $x$ and another data point $y\in\R^n$ is given as
    \begin{align*}
    d_1\big(\Phi(x),\Phi(y) \big) \coloneqq \big(\Phi_{j^*}(x)- \Phi_{j^*}(y) \big)^2.  
    \end{align*}
    On the other hand, the vector $[\Phi_j(x)]_{j =1}^m$ is usually normalized to a probability vector $[\Tilde{\Phi}_j(x)]_{j=1}^m$ by applying the softmax function, namely $\Tilde{\Phi}_j(x) \coloneqq \exp{\Phi_j(x)}/\sum_{i = 1}^m\exp{\Phi_i(x)}$. This, in turn, gives another measure of the distortion between $\Phi(x), \Phi(y) \in \R^m$, namely
    \begin{align*}
        d_2\big(\Phi(x),\Phi(y) \big) \coloneqq \big(\Tilde{\Phi}_{j^*}(x)- \Tilde{\Phi}_{j^*}(y) \big)^2,
    \end{align*}
    where $j^* \coloneqq \argmax_{j \in \set{1, \hdots, m}} \Phi_j(x) = \argmax_{j \in \set{1, \hdots, m}} \Tilde{\Phi}_j(x)$. 
An important property of the softmax function is the invariance under translation by a vector $[c,\hdots,c]^\top \in \R^m$, where $c\in\R$ is a constant.
%Thus, the probability assigned to each class is kept unchanged when the pre-softmax scores are shifted by the same amount.
By definition, only $d_2$ respects this invariance while $d_1$ does not. 
\end{example}

\begin{example}[Measure of distortion for regression task]\label{example:regression_measure}
    Consider a regression model $\Phi:\R^n\to\R^m$ and an input signal $x \in \R^n$. One can then define the measure of distortion between the outputs of $x$ and another data point $y\in\R^n$ as
    \begin{align*}
        d_3\big((\Phi(x),\Phi(y)\big) \coloneqq \norm{\Phi(x)- \Phi(y)}_2^2.
    \end{align*}
    Sometimes it is reasonable to consider a certain subset of components $J \subseteq \set{1,\hdots,m}$ of the output vectors instead of all $m$ entries. Denoting the vector formed by corresponding entries by $\Phi_J(x)$, the measure of distortion between the outputs can be defined as
    \begin{align*}
        d_4\big((\Phi(x),\Phi(y)\big) \coloneqq \norm{\Phi_J(x)- \Phi_J(y)}_2^2.
    \end{align*}
    The measure $d_4$ will be used in our experiments for radio maps in Subsection \ref{sec:radiomap}.
\end{example}

\subsection{Implementation}\label{section:implementation}

The RDE mask from \Cref{def:rde mask} was defined as a solution to
\begin{align*}
    \min_{s\in \{0,1\}^k} \quad D(x,s,\mathcal{V}_s, \Phi) \quad \text{ s.t. } \quad \norm{s}_0 \leq \ell.
\end{align*}
In practice, we need to relax this problem. We offer the following three approaches.

\subsubsection{$\ell_1$-relaxation with Lagrange multiplier.}\label{subsubsection:ell1-relaxation}
The RDE mask can be approximately computed by finding an approximate solution to the following relaxed minimization problem:
\begin{align}\label{eq:relax_problem_ell1}\tag{$\calP_1$}
    \min_{s\in [0,1]^k} \quad D(x,s,\mathcal{V}_s, \Phi) + \lambda \|s\|_1,
\end{align}
where $\lambda>0$ is a hyperparameter for the sparsity level. Note that the optimization problem is not necessarily convex, thus the solution might not be unique.

%The expected distortion $D(x,s,\mathcal{V}_s, \Phi)$ can be approximated with assumed density filtering as discussed in \cite{RDE_original_2019}. If the distribution $\calV_s$ is chosen in a way such that the samples are i.i.d., e.g., when the obfuscation strategy is blurring or injecting Gaussian noise, one can simply apply Monte-Carlo estimates.  
The expected distortion $D(x,s,\mathcal{V}_s, \Phi)$ can typically be approximated with simple Monte-Carlo estimates, i.e., by averaging i.i.d. samples from $\calV_s$.
After estimating $D(x,s,\mathcal{V}_s, \Phi)$, one can optimize the mask $s$ with stochastic gradient descent (SGD) to solve the optimization problem \cref{eq:relax_problem_ell1}. %For in-distribution interpretability the implementation becomes more involved since one needs to first train an in-painting GAN (or a different powerful conditional generative model). 

\subsubsection{Bernoulli relaxation.}
By viewing the binary mask as Bernoulli random variables $s\sim\text{Ber}(\theta)$ and optimizing over $\theta$, one can guarantee that the expected distortion $D(x,s,\mathcal{V}_s, \Phi)$ is evaluated on binary masks $s\in\{0,1\}^n$. To encourage sparsity of the resulting mask, one can still apply $\ell_1$-regularization on $s$, giving rise to the following optimization problem:
\begin{align}\label{eq:bernoulli_relax}\tag{$\calP_2$}
    \min_{\theta \in [0,1]^k}\quad  \mathop{\mathbb{E}}_{s\sim\text{Ber}(\theta)}\Bigg[D(x,s,\mathcal{V}_s, \Phi)\Bigg] + \lambda \norm{s}_1.
\end{align}
 Optimizing the parameter $\theta$ requires a continuous relaxation to apply SGD. This can be done using the concrete distribution \cite{maddison2017concrete}, which samples $s$ from a continuous relaxation of the Bernoulli distribution.
 
 %\ron{If I recall correctly, the masks where not simple random $\{0,1\}^d$ functions. Cosmas started with a Gaussian noise RDE, got masks from optimizing explanations, and then used these to train the inpainter, and so on. The initial GAn might have been with randomly generated masks. But I am not sure...}

\subsubsection{Matching pursuit.}\label{subsubsec: matching pursuit}
As an alternative, one can also perform matching pursuit \cite{matching_pursuit_1993}. Here, the non-zero entries of $s\in \{0,1\}^n$ are determined sequentially in a greedy fashion to minimize the resulting distortion in each step. More precisely, we start with a zero mask $s^0=0$ and gradually build  up the mask by updating $s^t$ at step $t$ by the rule given by
\begin{align*}
    s^{t+1} = s^t + \argmin_{e_j:\,s_j^t=0} \, D(x,s^{t}+e_j,\calV_s,\Phi).
\end{align*}
Here, the minimization is taken over all standard basis vectors $e_j \in \R^k$ with $s_j^t = 0$. The algorithm terminates when reaching some desired error tolerance or after a prefixed number of iterations. While this means that in each iteration we have to test every entry of $s$, it is applicable when $k$ is small or when we are only interested in very sparse masks. %\ron{It is not necessary. We can always take a non-binary mask and project it to a binary one.}
%\stefan{I changed the last sentence: before you said "While this means that in each iteration we have to test every entry of $s$, it is applicable in cases where k is small or  we are only interested in very highly sparse masks." I think its better to say it is neccessary when require binary masks, e.g. perturbations on edges in a graph. Do you want to name that as an example? } 

\section{Experiments} \label{sec:experiments}
%With our experiments, we demonstrate how capable the RDE framework is for analyzing different data modalities. Whereas most works in the field focus on images, we choose two challenging modalities that are often unexplored. The first is audio, where we focus on classification of acoustic instruments in the NSynth dataset \cite{engel2017neural}. In this setting, we train an inpainting network $G$ in order to inform $y$.

%The second is interpreting the outcome of physical simulations used to estimate radio maps in urban environments. In this setting, we take a different tack with our inpainter. Because the data is expensive to gather, highly structured, and has capable associated physical simulations, we rely on a model-based approach along with heuristics to in-paint. We optimize $s$ with MP as described in \Cref{section:implementation}. 
With our experiments, we demonstrate the broad applicability of the generalized RDE framework. Moreover, our experiments illustrate how different choices of obfuscation strategies, optimization procedures, measures of distortion, and input signal representations, discussed in \Cref{sec:general_formulation}, can be leveraged in practice. 
We explain model decisions on various challenging data modalities and tailor the input signal representation and measure of distortion to the domain and interpretation query. In \Cref{section:experiment_images}, we focus on image classification, a common baseline task in the interpretability literature. In Sections \ref{section:experiments_audio} and \ref{sec:radiomap}, we consider two other data modalities that are often unexplored. \Cref{section:experiments_audio} focuses on audio data, where the underlying task is to classify acoustic instruments based on a short audio sample of distinct notes, while in  \Cref{sec:radiomap}, the underlying task is a regression with data in the form of physical simulations in urban environments. We also believe
our explanation framework sustains applications beyond interpretability tasks. An example is given in \Cref{subsubsec:interpretation driven training}, where we add an RDE inspired regularizer to the training objective of a radio map estimation model.

\subsection{Images}\label{section:experiment_images}

We begin with the most ordinary domain in the interpretability literature: image classification tasks. The authors of \cite{RDE_original_2019} applied RDE to image data before by considering pixel-wise perturbations. We refer to this method as \emph{Pixel RDE}. Other explanation methods \cite{LIME_2016}, \cite{Layerwise_relevance_prop2015}, \cite{chang_GAN_explanation_2018}, and \cite{Dabkowski_Gal_2017}, have also previously exclusively operated in the pixel domain. In \cite{kolek2021cartoon}, we challenged this customary practice by successfully applying RDE in a wavelet basis, where sparsity translates into piece-wise smooth images (also called cartoon-like images). The novel explanation method was coined \emph{CartoonX} \cite{kolek2021cartoon} and extracts the relevant piece-wise smooth part of an image. First, we review the Pixel RDE method and present experiments on the ImageNet dataset \cite{imagenet_2009}, which is commonly considered a challenging classification task. Finally, we present CartoonX and discuss its advantages. For all the ImageNet experiments, we use the pre-trained \emph{MobileNetV3-Small} \cite{Howard2019SearchingFM}, which achieved a top-1 accuracy of 67.668\% and a top-5 accuracy of 87.402\%, as the classifier.  

\begin{figure}[h]
	\centering
	\begin{subfigure}{0.3\textwidth}\centering%no!\hfill
		\includegraphics[scale=0.25]{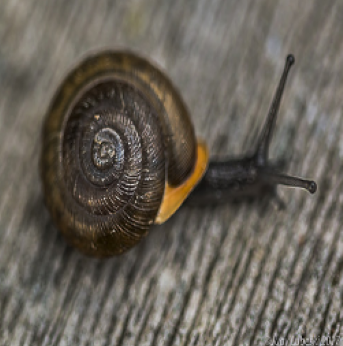}
		\caption{}
		\label{fig:pixel rde orig1}
	\end{subfigure}%
	\hfill
	\begin{subfigure}{0.4\textwidth}\centering%no!\hfill
		\includegraphics[scale=0.25]{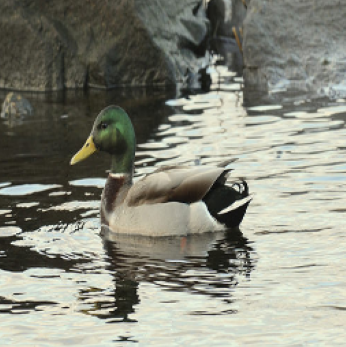}
		\caption{}
		\label{fig:pixel rde orig2}
	\end{subfigure}%
	\begin{subfigure}{0.3\textwidth}\centering%no!\hfill
		\includegraphics[scale=0.25]{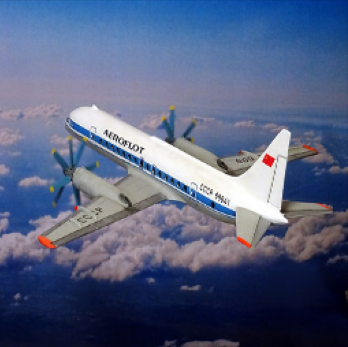}
		\caption{}
		\label{fig:pixel rde orig3}
	\end{subfigure}%
	\hfill
	\begin{subfigure}{0.3\textwidth}\centering
		\includegraphics[scale=0.25]{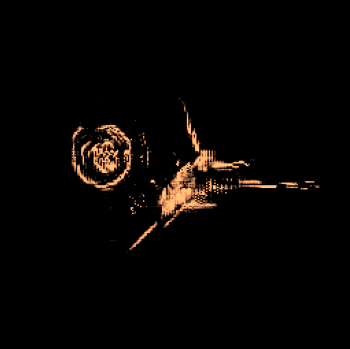}
		\caption{}
		\label{fig:pixel mask1}
	\end{subfigure}%
		\hfill
	\begin{subfigure}{0.3\textwidth}\centering
		\includegraphics[scale=0.25]{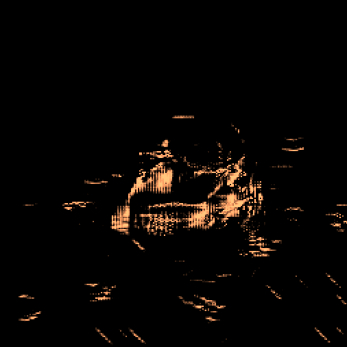}
		\caption{}
		\label{fig:pixel mask2}
	\end{subfigure}%
	    \hfill
	\begin{subfigure}{0.3\textwidth}\centering
		\includegraphics[scale=0.25]{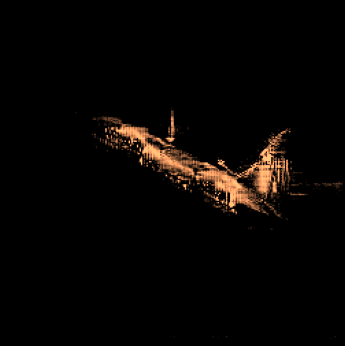}
		\caption{}
		\label{fig:pixel mask2}
	\end{subfigure}%
		\hfill
	\begin{subfigure}{0.3\textwidth}\centering
		\includegraphics[scale=0.25]{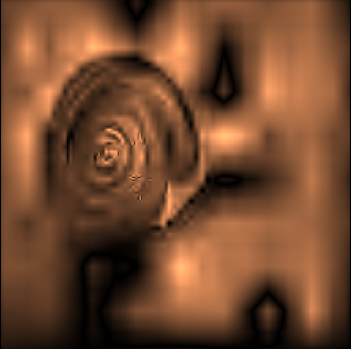}
		\caption{}
		\label{fig:greyinv1}
	\end{subfigure}%
		\hfill
	\begin{subfigure}{0.3\textwidth}\centering
		\includegraphics[scale=0.25]{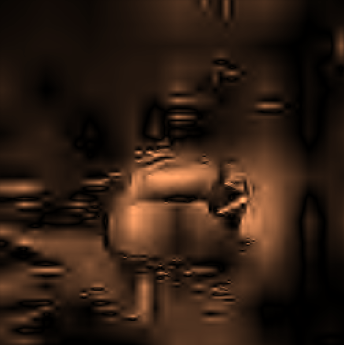}
		\caption{}
		\label{fig:greyinv2}
	\end{subfigure}%
	    \hfill
	\begin{subfigure}{0.3\textwidth}\centering
		\includegraphics[scale=0.25]{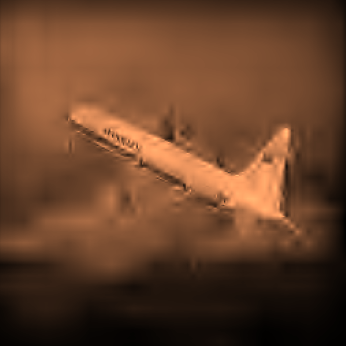}
		\caption{}
		\label{fig:greyinv3}
	\end{subfigure}%
	\caption{Top row: original images correctly classified as (a) snail, (b) male duck, and (c) airplane. Middle row: Pixel RDEs. Bottom row: CartoonX. Notably, CartoonX is roughly piece-wise smooth and overall more interpretable than the jittery Pixel RDEs.}
	\label{fig:image explanations}
\end{figure}
\subsubsection{Pixel RDE.}
Consider the following pixel-wise representation of an RGB image  $x\in \R^{3\times n}$:
$$
f: \prod_{i=1}^n \R^3 \to \R^{n\times 3},\; x = f(h_1,...,h_n),
$$
where $h_i\in\R^3$ represents the three color channel values of the $i$-th pixel in the image $x$, i.e. $(x_{i,j})_{j=1,..,3}=h_{i}$.
In pixel RDE a sparse mask $s\in [0,1]^n$ with $n$ entries---one for each pixel---is optimized to achieve low expected distortion $D(x,s,\mathcal{V}_s, \Phi)$. The obfuscation of an image $x$ with the pixel mask $s$ and a distribution $v\sim\mathcal{V}_s$ on $\prod_{i=1}^n \R^3$ is defined as $f(s \odot h + (1-s)\odot v)$.
In our experiments, we initialize the mask with ones, i.e., $s_i = 1$ for every $i \in \set{1,\hdots, n}$, and consider Gaussian noise perturbations $\mathcal{V}_s = \mathcal{N}(\mu,\Sigma)$. We set the noise mean $\mu\in\R^{3\times n}$ as the pixel value mean of the original image $x$ and the covariance matrix $\Sigma\coloneqq\sigma^2\Id\in\R^{3n\times 3n}$  as a diagonal matrix with $\sigma>0$ defined as the pixel value standard deviation of the original image $x$. We then optimize the pixel mask $s$ for 2000 gradient descent steps on the $\ell_1$-relaxation of the RDE objective (see Section \ref{subsubsection:ell1-relaxation}).
We computed the distortion $d\big(\Phi(x),\Phi(y) \big)$ in $D(x,s,\mathcal{V}_s, \Phi)$ in the post-softmax activation of the predicted label multiplied by a constant $C=100$, i.e., 
\begin{align*}
d\big(\Phi(x),\Phi(y) \big) \coloneqq C\big(\Phi_{j^*}(x)- \Phi_{j^*}(y) \big)^2.  
\end{align*}
 The expected distortion $D(x,s,\mathcal{V}_s, \Phi)$ was approximated as a simple Monte-Carlo estimate after sampling 64 noise perturbations. For the sparsity level, we set the Lagrange multiplier to $\lambda=0.6$. All images were resized to 256 $\times$ 256 pixels. The mask was optimized for 2000 steps using the Adam optimizer with step size $0.003$. In the middle row of Figure \ref{fig:image explanations}, we show three example explanations with Pixel RDE for an image of a snail, a male duck, and an airplane, all from the ImageNet dataset. Pixel RDE highlights as relevant both the snail's inner shell and part of its head, the lower segment of the male duck along with various lines in the water, and the airplane's fuselage and part of its rudder.

\subsubsection{CartoonX.}
\begin{figure}[h]
	\centering
	\begin{subfigure}{0.5\textwidth}\centering%no!\hfill
		\includegraphics[scale=0.5]{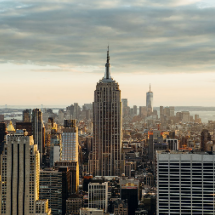}
		\caption{}
		\label{fig:dwt example1}
	\end{subfigure}%
	\hfill
	\begin{subfigure}{0.5\textwidth}\centering%no!\hfill
		\includegraphics[scale=0.5]{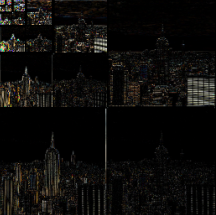}
		\caption{}
		\label{fig:dwt example2}
	\end{subfigure}%

	\caption{Discrete Wavelet Transform of an image: (a) original image (b) discrete wavelet transform. The coefficients of the largest quadrant in (b) correspond to the lowest scale and coefficients of smaller quadrants gradually build up to the highest scales, which are located in the four smallest quadrants. Three nested L-shaped quadrants represent horizontal, vertical and diagonal edges at a resolution determined by the associated scale. }
	\label{fig:dwt example}
\end{figure}

% Introduce wavelet representation of image
Formally, we represent an RGB image $x\in [0,1]^{3\times n}$ in its wavelet coefficients $h = \{h_i\}_{i=1}^n \in \prod_{i=1}^n\R^3$  with $J \in \set{{1, \hdots, \lfloor \log_2 n \rfloor}}$ scales as 
$
x = f(h)
$,
where f is the discrete inverse wavelet transform. Each $h_i = (h_{i,c})_{c=1}^3\subseteq \R^3$ contains three wavelet coefficients of the image, one for each color channel and is associated with a scale $k_i\in \set{1, \hdots, J}$ and a position in the image. Low scales describe high frequencies and high scales describe low frequencies at the respective image position.
We briefly illustrate  the wavelet coefficients in Figure \ref{fig:dwt example}, which visualizes the discrete wavelet transform of an image. 

\begin{figure}[h]
	\centering
	\begin{subfigure}{0.3\textwidth}\centering%no!\hfill
		\includegraphics[width=0.997\linewidth]{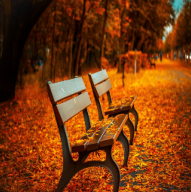}
		\caption{}
		\label{fig:orig}
	\end{subfigure}%
	\hfill
	\begin{subfigure}{0.3\textwidth}\centering
		\includegraphics[width=0.997\linewidth]{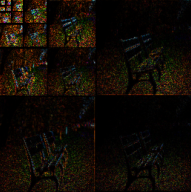}
		\caption{}
		\label{fig:dwt}
	\end{subfigure}%
	\hfill
	\begin{subfigure}{0.3\textwidth}\centering
		\includegraphics[width=0.997\linewidth]{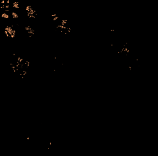}
		\caption{}
	   \label{fig:dwt mask}
	\end{subfigure}%
		\hfill
	\begin{subfigure}{0.3\textwidth}\centering
		\includegraphics[width=0.997\linewidth]{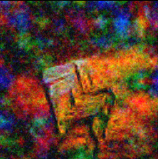}
		\caption{}
	   \label{fig:noise inv}
	\end{subfigure}%
		\hfill
	\begin{subfigure}{0.3\textwidth}\centering
		\includegraphics[width=0.997\linewidth]{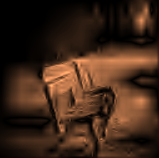}
		\caption{}
	   \label{fig:grey inv}
	\end{subfigure}%
	    \hfill
	\begin{subfigure}{0.3\textwidth}\centering
		\includegraphics[width=0.997\linewidth]{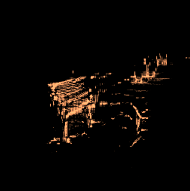}
		\caption{}
	   \label{fig:pixel mask}
	\end{subfigure}%
	\caption{CartoonX machinery: (a) image classified as park-bench, (b) discrete wavelet transform of the image, (c) final mask on the wavelet coefficients after the RDE optimization procedure, (d) obfuscation with final wavelet mask and noise, (e) final CartoonX, (f) Pixel RDE for comparison.}
	\label{fig:wavelet RDE comparison}
\end{figure}

CartoonX \cite{kolek2021cartoon} is a special case of the generalized RDE framework, particularly a special case of Example \ref{ex:FFT}, and optimizes a sparse mask $s\in[0,1]^n$ on the wavelet coefficients (see Figure \ref{fig:dwt mask}) so that the expected distortion $D(x,s,\mathcal{V}_s, \Phi)$ remains small. The obfuscation of an image $x$ with a wavelet mask $s$ and a distribution $v\sim\mathcal{V}_s$ on the wavelet coefficients is  $f(s \odot h  + (1-s)\odot v)$.
In our experiments, we used Gaussian noise perturbations and chose the standard deviation and mean adaptively for each scale: the standard deviation and mean for wavelet coefficients of scale $j\in\set{1, \hdots, J}$ were chosen as the standard deviation and mean of the wavelet coefficients of scale $j\in\set{1, \hdots, J}$ of the original image. 
Figure \ref{fig:noise inv} shows the obfuscation $f(s \odot h  + (1-s)\odot v)$ with the final  wavelet mask $s$ after the RDE optimization procedure. In Pixel RDE, the mask itself is the explanation as it lies in pixel space (see middle row in Figure \ref{fig:image explanations}), whereas the CartoonX mask lies in the wavelet domain. To go back to the natural image domain, we multiply the wavelet mask element-wise with the wavelet coefficients of the original greyscale image and invert this product back to pixel space with the discrete inverse wavelet transform.
The inversion is finally clipped into $[0,1]$ as are obfuscations during the RDE optimization to avoid overflow (we assume here the pixel values in $x$ are normalized into $[0,1]$). The clipped inversion in pixel space is the final CartoonX explanation (see Figure \ref{fig:grey inv}).

\begin{figure}
    \begin{center}
        \scalebox{0.6}{\input{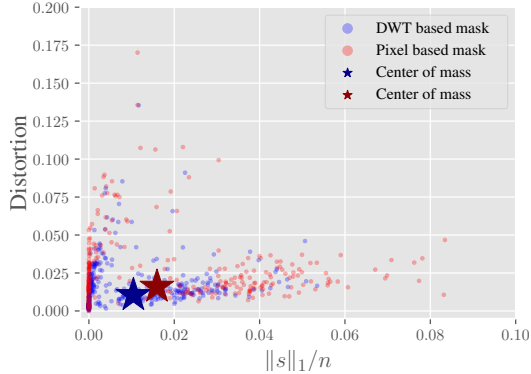}}
    \end{center}
    \caption{Scatter plot of rate-distortion in pixel basis and wavelet basis. Each point is an explanation of a distinct image in the ImageNet dataset with  distortion and normalized $\ell_1$-norm measured for the final mask. The wavelet mask achieves lower distortion than the pixel mask, while using less coefficients.}
    % In the scatter plot we used lambda 2.5 for both pixel and wavelet RDE
    \label{fig:scatter plot}
\end{figure}

The following points should be kept in mind when interpreting the final CartoonX explanation, i.e., the inversion of the wavelet coefficient mask: (1) CartoonX provides the relevant pice-wise smooth part of the image. (2) The inversion of the wavelet coefficient mask was not optimized to be sparse in pixel space but in the wavelet basis. (3) A region that is black in the inversion could nevertheless be relevant if it was already black in the original image. This is due to the multiplication of the mask with the wavelet coefficients of the greyscale image before taking the discrete inverse wavelet transform. (4) Bright high resolution regions are relevant in high resolution and bright low resolution regions are relevant in low resolution. (5) It is inexpensive for CartoonX to mark large regions in low resolution as relevant. (6) It is expensive for CartoonX to mark large regions in high resolution as relevant.

In \Cref{fig:image explanations}, we compare CartoonX to Pixel RDE. The piece-wise smooth wavelet explanations are more interpretable than the jittery Pixel RDEs. In particular, CartoonX asserts that the snail's shell without the head suffices for the classification, unlike Pixel RDE, which insinuated that both the inner shell and part of the head are relevant. Moreover, CartoonX shows that the water gives the classifier context for the classification of the duck, which one could have only guessed from the Pixel RDE. Both Pixel RDE and CartoonX state that the head of the duck is not relevant. Lastly, CartoonX, like Pixel RDE, confirms that the wings play a subordinate role in the classification of the airplane.

\subsubsection{Why explain in the wavelet basis?} 
%\ron{Do you need a question mark in the title? Or perhaps: wavelet vs. pixel explanations}\stefan{I would try with a question mark}
Wavelets provide optimal representation for piece-wise smooth 1D functions \cite{devore_1998}, and represent 2D piece-wise smooth images, also called \emph{cartoon-like images} \cite{kutyniok2010}, efficiently as well \cite{Romberg06wavelet-domainapproximation}. Indeed, sparse vectors in the wavelet coefficient space encode cartoon-like images reasonably well \cite{STEPHANE2009535}, certainly better than sparse pixel representations.
Moreover, the optimization process underlying CartoonX produces sparse vectors in the wavelet coefficient space. Hence CartoonX typically generates cartoon-like images as explanations. This is the fundamental difference to Pixel RDE, which produces rough, jittery, and pixel-sparse explanations. Cartoon-like images are more interpretable and provide a natural model of simplified images.
Since the goal of the RDE explanation is to generate an easy to interpret simplified version of the input signal, we argue that CartoonX explanations are more appropriate for image classification than Pixel RDEs. 
Our experiments confirm that the CartoonX explanations are roughly piece-wise  smooth  explanations and are overall more interpretable than Pixel RDEs (see Figure \ref{fig:image explanations}).

\subsubsection{CartoonX implementation.}
% Explain implementation
Throughout our CartoonX experiments we chose the Daubechies 3 wavelet system, $J=5$ levels of scales and zero padding for the discrete wavelet transform. For the implementation of the discrete wavelet transform, we used the Pytorch Wavelets package, which supports gradient computation in Pytorch. 
Distortion was computed as in the Pixel RDE experiments. The perturbations $v\sim\mathcal{V}_s$ on the wavelet coefficients were chosen as Gaussian noise with standard deviation and mean computed adaptively per scale. As in the Pixel RDE experiments, the wavelet mask was optimized for 2000 steps with the Adam optimizer to minimize the $\ell_1$-relaxation of the RDE objective. We used $\lambda=3$ for CartoonX.
\subsubsection{Efficiency of CartoonX.}
Finally, we compare Pixel RDE to CartoonX quantitatively by analyzing the distortion and sparsity associated with the final explanation mask. Intuitively, we expect the CartoonX method to have an efficiency advantage, since the discrete wavelet transform already encodes natural images sparsely, and hence less wavelet coefficients are required to represent images than pixel coefficients.
%\ron{more accurately: since the DWT already encodes natural images sparsely, and hence less wavelet coefficients are required to represent images than pixel coefficients.}.
Our experiments confirmed this intuition, as can be seen in the scatter plot in Figure \ref{fig:scatter plot}.

\subsection{Audio} \label{section:experiments_audio}
We consider the NSynth dataset \cite{engel2017neural}, a library of short audio samples of distinct notes played on a variety of instruments. We pre-process the data by computing the power-normalized magnitude spectrum and phase information using the discrete Fourier transform on a logarithmic scale from $20$ to $8000$ Hertz. Each data instance is then represented by the magnitude and the phase of its Fourier coefficients as well as the discrete inverse Fourier transform (see \Cref{ex:FFT}).

\subsubsection{Explaining the classifier.}
Our model $\Phi$ is a network trained to classify acoustic instruments. We compute the distortion with respect to the pre-softmax scores, i.e., deploy $d_1$ in \Cref{example:classification_measure} as the measure of distortion. We follow the obfuscation strategy described in \Cref{ex:GAN_obfuscation} and train an inpainter $G$ to generate the obfuscation $G(h,s,z)$. Here, $h$ corresponds to the representation of a signal, $s$ is a binary mask and $z$ is a normally distributed seed to the generator. 

\begin{figure}[h]
    \centering
    \input{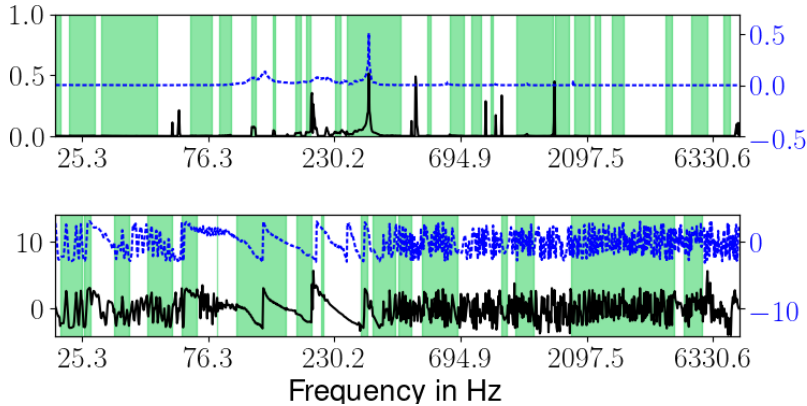}
    \caption{{Inpainted Bass}: Example inpainting from $G$. The bottom plot depicts phase versus frequency and the top plot depicts magnitude versus frequency. The random binary mask is represented by the green parts. The axes for the inpainted signal (black) and the original signal (blue dashed) are offset to improve visibility. Note how the inpainter generates plausible peaks in the magnitude and phase spectra, especially with regard to rapid ($\ge 600$Hz) versus smooth ($< 270$Hz) changes in phase.}
    % \caption{Example output of the inpainter $G$ for a randomly generated mask (indicated by the green regions). The axes for the inpainted signal (black) and the original signal (blue dashed) are offset to improve visibility. Note how the inpainter generates plausible peaks in the magnitude spectrum and infills the phase spectrum plausibly with regard to rapid (seen above 600Hz) vs. smooth (seen between 70Hz and 270Hz) change in phase angle.}
    \label{fig:gan_output}
\end{figure}

We use a residual CNN architecture for $G$ with added noise in the input and deep features. More details can be found in Section \ref{subsubsec:architecture inpainting}. We train $G$ until the outputs are found to be satisfactory, exemplified by the outputs in \Cref{fig:gan_output}.

\begin{figure}[h]
	\centering
	\begin{subfigure}{0.9\textwidth}\centering%no!\hfill
		\includegraphics[width=0.7\linewidth]{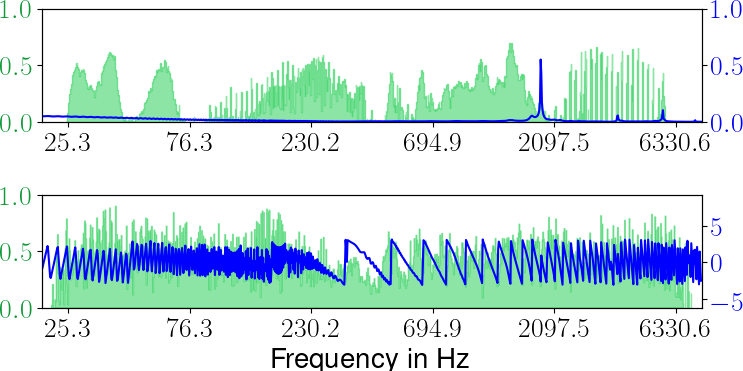}\\
	    \caption{Guitar}
	    \label{fig:nsynth_exp1}
	\end{subfigure}%
	\vspace{0.5cm}
	\begin{subfigure}{0.9\textwidth}\centering
	    \includegraphics[width=0.7\linewidth]{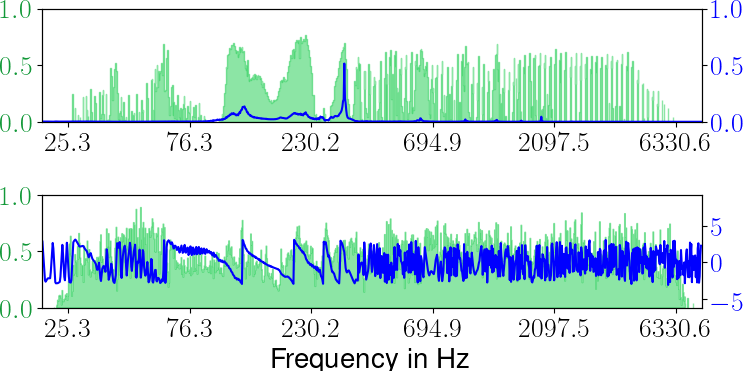}
	    \caption{Bass}
	    \label{fig:nsynth_exp2}
	\end{subfigure}%
	\caption{{Interpreting NSynth Model}: The optimized importance parameter $\theta$ (green) overlayed on top of the DFT (blue). For each of guitar and bass, the top graph shows the power-normalized magnitude and the bottom the phase. Notice the solid peaks between 30Hz and 60Hz for guitar and between 100Hz and 230Hz for bass. These occur because the model is relying on those parts of the spectra, for the classification. Notice also how many parts of the spectrum are important even when the magnitude is near zero. This indicates that the model pays attention to whether those frequencies are missing.}
    % \caption{Graphical results of what is important in two NSynth classification examples. The importance is given by the optimized parameter $\theta$ shown in green. Overlayed is the signal from the DFT of the audio sample (blue). For each guitar and bass the top graph shows the magnitude and the bottom graph the phase angle for each frequency. Notice the solid peaks between 30Hz and 60Hz for the guitar classification and the peaks between 100Hz and 230Hz for the bass classification. Notice also how many parts of the spectrum are important even if the magnitude is near zero. This indicates that there is a need for these frequencies to be \emph{missing} for the classification to be retained.}    
    \label{fig:nsynth_explainability}
\end{figure}

To compute the explanation maps, we numerically solve \Cref{eq:bernoulli_relax} as discussed in Subsection \ref{section:implementation}. In particular, $s$ is a binary mask indicating whether the phase and magnitude information of a certain frequency should be dropped and is specified as a Bernoulli variable $s \sim \text{Ber}(\theta)$. We chose a regularization parameter of $\lambda = 50$ and minimized the corresponding objective using the Adam optimizer with a step size of $10^{-5}$ in $10^6$ iterations. For the concrete distribution, we used a temperature of $0.1$. Two examples resulting from this process can be seen in \Cref{fig:nsynth_explainability}.

Notice here that the method actually shows a strong reliance of the classifier on low frequencies (30Hz-60Hz) to classify the top sample in \Cref{fig:nsynth_explainability} as a guitar, as only the guitar samples have this low frequency slope in the spectrum. We can also see in contrast that classifying the bass sample relies more on the continuous signal between 100Hz and 230Hz. %\stefan{Regarding the phase, it is interesting that if the phase angle changes smoothly with frequency, the model pays less attention than if the phase angle changes rapidly. This can also be explained by only needing fewer samples throughout different frequencies to recognize smooth phase angle changes (versus rapid ones).} \ron{This may also be an artifact of the GAN inpainter, that only needs a few samples throughout different frequencies to inpaint smooth phase angle changes (versus rapid ones).}

%\ron{This can be explained by the fact that the inpainter can leave big gaps throughout different frequencies and still recognize that the phase should be smooth there. The inpainter can then inpaint the missing information quite accurately, so it can be left out of teh mask with relatively low degradation to the accuracy.
%This may also be an artifact of the GAN inpainter, that only needs a few samples throughout different frequencies to inpaint smooth phase angle changes (versus rapid ones).}\stefan{we might need to reformulate "leave big gaps" since a reader might not understand what we mean.} \ron{Perhaps we leave this whole discussion out? It requires too much discussion about stuff we did not address up until now, like, some masks can be artifacts of the GAN.}\stefan{I commented out the last two lines in the text above.}

\subsubsection{Magnitude vs Phase.}

In the above experiment, we have represented the signals by the magnitude and phase information at each frequency, hence the mask $s$ acts on each frequency. Now we consider the \emph{interpretation query} of whether the entire magnitude spectrum or the entire phase spectrum is more relevant for the prediction. Accordingly, we consider the representation discussed in \Cref{ex:query_phase_magnitude} and apply the mask $s$ to turn off or on the whole magnitude spectrum or the phase information. Furthermore, we can optimize $s$ not only for one datum but for all samples from a class. This extracts the information whether magnitude or phase is more important for predicting samples from a specific class.

For this, we again minimized \Cref{eq:bernoulli_relax} (meaned over all samples of a class) with $\theta$ as the Bernoulli parameter using the Adam optimizer for $2 \times 10^5$ iterations with a step size of $10^{-4}$ and the regularization parameter $\lambda=30$. Again, a temperature of $t=0.1$ was used for the concrete distribution. 

\begin{table}[h]
    %\label{sample-table}

    \begin{center}
    \begin{footnotesize}
    \begin{sc}
    \begin{tabular}{lrr}
    \toprule
    Intrument & Magnitude & Phase \\
     &  Importance &  Importance\\
    \midrule
    Organ  & 0.829 & 1.0\\
    Guitar & 0.0 & 0.999\\
    Flute  & 0.092 & 1.0\\
    Bass   & 1.0 & 1.0\\
    Reed   & 0.136 & 1.0\\
    Vocal  & 1.0 & 1.0\\
    Mallet & 0.005   & 0.217\\
    Brass  & 0.999 & 1.0\\
    Keyboard & 0.003 & 1.0\\
    String & 1.0 & 0.0\\
    \bottomrule
    \end{tabular}
    \end{sc}
    \end{footnotesize}
    \end{center}
    \caption{Magnitude importance versus phase importance.}
    \label{tab:mag_vs_phase}

\end{table}

From the results of these computations, which can be seen in \Cref{tab:mag_vs_phase}, we can observe that there is a clear difference on what the classifier bases its decision on across instruments. The classification of most instruments is largely based on phase information. For the mallet, the values are low for magnitude and phase, which means that the expected distortion is very low compared to the $\ell_1$-norm of the mask, even when the signal is completely inpainted. 
This underlines that the regularization parameter $\lambda$ may have to be adjusted for different data instances, especially when measuring distortion in the pre-softmax scores.

\subsubsection{Architecture of the inpainting network $G$.}\label{subsubsec:architecture inpainting}

\begin{figure}[hb]
    \begin{center}
    \input{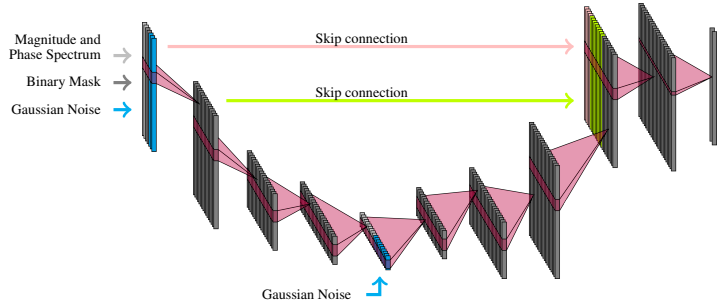}
    \caption{Diagram of the inpainting network for NSynth.}
    \label{fig:inpainter_diagram}
    \end{center}
\end{figure}

Here, we briefly describe the architecture of the inpainting network $G$ that was used to generate obfuscations to the target signals. In particular, \Cref{fig:inpainter_diagram} shows the diagram of the network $G$ and \Cref{tab:inpainter_layers} shows information about its layers. 
%\ron{write something here...}

\begin{table*}[h]
\label{sample-table}
\vskip 0.15in
\begin{center}
\begin{small}
\begin{sc}
\begin{tabular}{lrrr}
\toprule
Layer & Filter Size & Output Shape & \# Params\\
\midrule
Conv1d-1        & 21  & [-1, 32, 1024] & 4,736\\
ReLU-2          &     & [-1, 32, 1024] & 0\\
Conv1d-3        & 21  & [-1, 64, 502]  & 43,072\\
ReLU-4          &     & [-1, 64, 502]  & 0\\
BatchNorm1d-5   &     & [-1, 64, 502]  & 128\\
Conv1d-6        & 21  & [-1, 128, 241] & 172,160\\
ReLU-7          &     & [-1, 128, 241] & 0\\
BatchNorm1d-8   &     & [-1, 128, 241] & 256\\
Conv1d-9        & 21  & [-1, 16, 112]  & 43,024\\
ReLU-10         &     & [-1, 16, 112]  & 0\\
BatchNorm1d-11  &     & [-1, 16, 112]  & 32\\
ConvTranspose1d-12& 21& [-1, 64, 243]  & 43,072\\
ReLU-13         &     & [-1, 64, 243]  & 0\\
BatchNorm1d-14  &     & [-1, 64, 243]  & 128\\
ConvTranspose1d-15& 21& [-1, 128, 505] & 172,160\\
ReLU-16         &     & [-1, 128, 505] & 0\\
BatchNorm1d-17  &     & [-1, 128, 505] & 256\\
ConvTranspose1d-18& 20& [-1, 64, 1024] & 163,904\\
ReLU-19         &     & [-1, 64, 1024] & 0\\
BatchNorm1d-20  &     & [-1, 64, 1024] & 128\\
Skip Connection &     & [-1, 103, 1024]& 0\\
Conv1d-21       & 7   & [-1, 128, 1024]& 92,416\\
ReLU-22         &     & [-1, 128, 1024]& 0 \\
Conv1d-23       & 7   & [-1, 2, 1024]  & 1,794\\
ReLU-24         &     & [-1, 2, 1024]  & 0\\
\bottomrule
Total number of parameters: & & & 737,266\\
\end{tabular}
\end{sc}
\end{small}
\end{center}
\caption{Layer table of the Inpainting model for the NSynth task.}
\label{tab:inpainter_layers}
\vskip -0.2in
\end{table*}

\subsection{Radio Maps}\label{sec:radiomap}

In this subsection, we assume a set of transmitting devices (Tx) broadcasting a signal within a city. The received strength varies with location and depends on physical factors such as line of sight, reflection, and diffraction. We consider the regression problem of estimating a function that assigns the proper signal strength to each location in the city. Our dataset $\mathcal{D}$ is RadioMapSeer~\cite{levie2019radiounet} containing 700 maps, 80 Tx per map, and a corresponding grayscale label encoding the signal strength at every location. Our model $\Phi$ receives as input $x = [x^{(0)},x^{(1)},x^{(2)}]$, where $x^{(0)}$ is a binary map of the Tx locations, $x^{(1)}$ is a noisy binary map of the city (where a few buildings are missing), and $x^{(2)}$ is a grayscale image representing a number of ground truth measurements of the strength of the signal at the measured locations and zero elsewhere. We apply the UNet \cite{UNet15,levie2019radiounet,levieIEEE2020} architecture and train $\Phi$ to output the estimation of the signal strength throughout the city that interpolates the input measurements.

\begin{figure}[h]
	\centering
	\begin{subfigure}{0.3\textwidth}\centering%no!\hfill
		\includegraphics[width=0.997\linewidth]{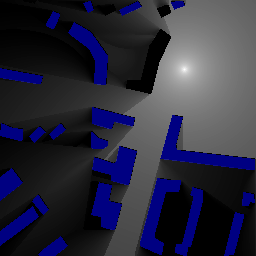}
		\caption{Ground Truth}
		\label{fig:radio1}
	\end{subfigure}%
	\hfill
	\begin{subfigure}{0.3\textwidth}\centering
		\includegraphics[width=0.997\linewidth]{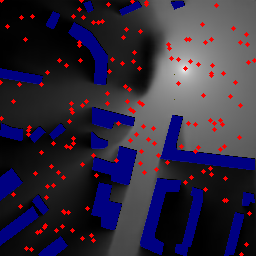}
		\caption{$\Phi$ Estimation}
		\label{fig:radio2}
	\end{subfigure}%
	\hfill
	\begin{subfigure}{0.3\textwidth}\centering
		\includegraphics[width=0.997\linewidth]{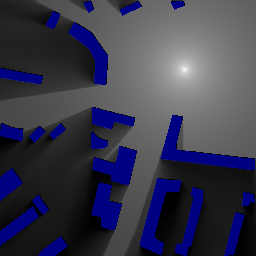}
		\caption{$\Phi_0$ Estimation}
	   \label{fig:radio3}
	\end{subfigure}%
	\caption{{Radio map estimations}: The radio map (gray), input buildings (blue), and input measurements (red).}
	\label{fig:radio}
\end{figure}

Apart from the model $\Phi$, we also have a simpler model $\Phi_0$%\ron{Perhaps we need to change the name $\phi_{gt}$ since this is not the ground truth. Call it $\Phi_0$ perhaps.}
, which only receives the city map and the Tx locations as inputs and is trained with unperturbed input city maps. This second model $\Phi_0$ will be deployed to inpaint measurements to input to $\Phi$. 
%We also consider a second model $\Phi_{\text{gt}}$ similar to the first except that it receives as input the ground truth city map along with the Tx locations. 
See \Cref{fig:radio1}, \ref{fig:radio2}, and \ref{fig:radio3} for examples of a ground truth map and estimations for $\Phi$ and $\Phi_{0}$, respectively.

 %We receive as input a perturbed map of the city, meaning that a few buildings are missing, along with some measurements of the target radio map at a few locations.
 %The model, called RadioUNet$_S$ ($S$ for samples), is trained to interpolate the measurements and use the perturbed city map to guide the output to be physically feasible. 
 %RadioUNet$_S$ is a CNN with UNet architecture that receives as inputs three channels, the binary map of the city, the binary map of the transmitter's location, and the measurements as a gray level image with the strength of the signal at the measured locations and zero elsewhere. RadioUNet$_S$ then outputs the estimation of the signal strength throughout the map. Apart from this model, we also consider a second, simpler model from \cite{levie2019radiounet}, called RadioUNet$_C$ ($C$ for clean), which only receives the city map and transmitter location as inputs and is trained with unperturbed input city maps. For an example of a ground truth radio map and its estimations via RadioUNet$_S$ and RadioUNet$_C$ see Figures \ref{fig:radio1}, \ref{fig:radio2}, and \ref{fig:radio3} respectively.

\subsubsection{Explaining Radio Map $\Phi$.}

Observe that in \Cref{fig:radio1} there is a missing building in the input (the black one) and in \Cref{fig:radio2}, $\Phi$ in-fills %\ron{in-fills/inpaints} 
this building with a shadow. As a black box method, it is unclear why it made this decision. Did it rely on signal measurements or on building patterns? 
To address this, we 
consider each building as a cluster of pixels and each measurement as potential targets for our mask $s = [s^{(1)}, s^{(2)}]$, where $s^{(1)}$ acts on buildings and $s^{(2)}$ acts on measurements. We then apply matching pursuit (see Subsection \ref{subsubsec: matching pursuit}) to find a minimal mask $s$ of critical components (buildings and measurements). 

To be precise, suppose we are given a target input signal $x = [x^{(0)}, x^{(1)}, x^{(2)}]$. Let $k_1$ denote the number of buildings in $x^{(1)}$ and $k_2$ denote the number of measurements in $x^{(2)}$. Consider the function $f_1$ that takes as inputs vectors in $\set{0,1}^{k_1}$, which indicate the existence of buildings in $x^{(1)}$, and maps them to the corresponding city map in the original city map format. Analogously, consider the  function $f_2$ that takes as input  the measurements in $\R^{k_2}$ and maps them to the corresponding grayscale image of the original measurements format. Then, $f_1$ and $f_2$ encode the locations of the buildings and measurements in the target signal $x=[x^{(0)}, f_1(h^{(1)}), f_2(h^{(2)})]$, where $h^{(1)}$ and $h^{(2)}$ denotes the building  and measurement representation of $x$ in $f_1$ and $f_2$. When $s^{(1)}$ has a zero entry, i.e., a building in $h^{(1)}$ was not selected, we replace the value in the obfuscation with zero (this corresponds to a constant perturbation equal to zero).  
Then, the obfuscation of the target signal $x$ with a mask $s=[s^{(1)}, s^{(2)}]$ and perturbations $v=[v^{(1)}, v^{(2)}]\coloneqq [0, v^{(2)}] $ becomes:
$$
y \coloneqq[x^{(0)}, f_1(s^{(1)}\odot h^{(1)}), f_2(s^{(2)}\odot h^{(2)}+ (1-s^{(2)})\odot v^{(2)})].
$$

%To be precise, suppose we are given a target input signal $x = [x^{(0)}, x^{(1)}, x^{(2)}]$, we will now describe a representation according to which we can then define obfuscations to $x$. Let $k_1$ denote the number of buildings in $x^{(1)}$ and $k_2$ denote the number of measurements in $x^{(2)}$. We define a function $f_1$ that takes inputs as vectors in $\set{0,1}^{k_1}$ indicating the existence of buildings in $x^{(1)}$ and maps it to the corresponding city map of the true format, as well as a function $f_2$ that takes inputs as the measurements in $\R^{k_2}$ and also maps it to the corresponding grayscale image of the true format. In particular, $f_1$ and $f_2$ encode the locations of the buildings and measurements in the target signal $x$ and the signals locally around $x$ are represented using the function $f \coloneqq [x_0, f_1(\cdot), f_2(\cdot)]$. At $x$, we have $x = f (h^{(1)}, h^{(2)}) = [x^{(0)}, f_1(h^{(1)}), f_2(h^{(2)})]$ where $h^{(1)} \coloneqq [1,\hdots,1]$ while $h^{(2)}$ consists the ground truth measurements sampled from $x^{(2)}$. According to this representation, the masks $s^{(1)}$ and $s^{(2)}$ have the same dimensions as $h^{(1)}$ and $h^{(2)}$ respectively, meaning that $s^{(1)} \in \set{0,1}^{k_1}$ and $s^{(2)} \in \set{0,1}^{k_2}$. We then define the obfuscation of the signal to be 
%\begin{align*}
%   y &\coloneqq f(s^{(1)} \odot h^{(1)}, s^{(2)} \odot h^{(2)})  \\
%    &= [x^{(0)}, f_1(s^{(1)}), f_2(s^{(2)} \odot h^{(2)} + (1-s^{(2)}) \odot v)]
%\end{align*}
%where $v$ denotes the measurements chosen by some obfuscation strategy. 

While it is natural to model masking out a building by simply zeroing out the corresponding cluster of pixels by choosing $v^{(1)}=0$, we need to also properly choose $v^{(2)}$ for the entries, where the mask $s^{(2)}$ takes value $0$, in order to obtain appropriate obfuscations. For this, we can deploy the second model $\Phi_0$ as an inpainter. We consider the following two extreme obfuscation strategies. 
%While it is natural to model masking out a building using $s^{(1)}$ by zeroing out the corresponding cluster of pixels, we need to assign real values to the measurements masked out by $s^{(2)}$ in order to obtain appropriate obfuscations. In other words, we need to properly inform $v$ at the entries where the mask $s^{(2)}$ takes value $0$. For this we can deploy the second model $\Phi_{\text{gt}}$ as an inpainter, in particular we consider the following two extreme obfuscation strategies. 
The first is to set also $v^{(2)}$ to zero, i.e., simply remove the unchosen measurements from the input, with the underlying assumption being that any subset of measurements is valid for a city map. In the other extreme case, we inpaint all unchosen measurements by sampling at their locations the estimated radio map obtained by $\Phi_0$ based on the buildings selected by $s^{(1)}$.
%The first is to keep the subset of measurements chosen by $s$ unchanged while setting the unchosen measurements simply to zero, with the underlying assumption being that any subset of measurements and buildings is valid for a city map. 

The two extreme measurement completion methods correspond to two extremes of the interpretation query. Filling-in the missing measurements by $\Phi_0$ tends to overestimate the strength of the signal because there are fewer buildings to obstruct the transmissions. The empty mask will complete all measurements to the maximal possible signal strength -- the free space radio map. The overestimation in signal strength is reduced when more measurements and buildings are chosen, resulting in darker estimated radio maps. Thus, this strategy is related to the query of which measurements and buildings are important to darken the free space radio map, turning it to the radio map produced by $\Phi$. In the other extreme, adding more measurements to the mask with a fixed set of buildings typically brightens the resulting radio map. This allows us to answer which measurements are most important for brightening the radio map. 

\begin{figure}[h]
	\centering
	\begin{subfigure}{0.3\textwidth}\centering%no!\hfill
		\includegraphics[width=0.997\linewidth]{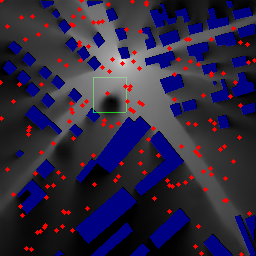}
		\caption{Estimated map. \newline \newline}
		\label{fig:radio_queries1}
	\end{subfigure}%
	\hfill
	\begin{subfigure}{0.3\textwidth}\centering
		\includegraphics[width=0.997\linewidth]{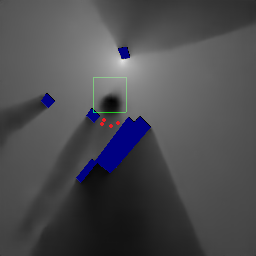}
		\caption{Explanation: Inpaint all unchosen measurements.}
		\label{fig:radio_queries2}
	\end{subfigure}%
	\hfill
	\begin{subfigure}{0.3\textwidth}\centering
		\includegraphics[width=0.997\linewidth]{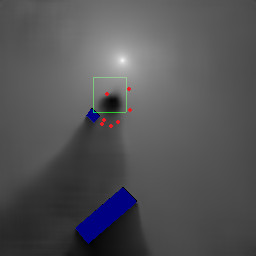}
		\caption{Explanation: Inpaint 2.5\% of unchosen measurements.}
	   \label{fig:radio_queries3}
	\end{subfigure}%
	\caption{{Radio map queries and explanations}: The radio map (gray), input buildings (blue), input measurements (red), and area of interest (green box). Middle represents the query ``How to fill in the image with shadows", while right is the query ``How to fill in the image both with shadows and bright spots?". We inpaint with $\Phi_0$.}
	\label{fig:radio_queries}
\end{figure}

Between these two extreme strategies lies a continuum of completion methods where a random subset of the unchosen measurements is sampled from $\Phi_0$, while the rest are set to zero. Examples of explanations of a prediction $\Phi(x)$ according to these methods are presented in \Cref{fig:radio_queries}. Since we only care about specific small patches exemplified by the green boxes, the distortion here is measured with respect to the $\ell_2$ distance between the output images restricted to the corresponding region (see also \Cref{example:regression_measure}).

 When the query is how to darken the free space radio map (\Cref{fig:radio_queries2}), the optimized mask $s$ suggests that samples in the shadow of the missing building are the most influential in the prediction. %.\cinjon{Why? What in the Figures says this? There's no explanation for either case as far as I can tell.}
These dark measurements are supposed to be in line-of-sight of a Tx, which indicates that the network deduced that there is a missing building. When the query is how to fill in the image both with shadows and bright spots (\Cref{fig:radio_queries3}), both samples in the shadow of the missing building and samples right before the building are influential. This indicates that the network used the bright measurements in line-of-sight and avoided predicting an overly large building. To understand the chosen buildings, note that $\Phi$ is based on a composition of UNets and is thus interpreted as a procedure of extracting high level and global information from the inputs to synthesize the output. The locations of the chosen buildings in \Cref{fig:radio_queries} reflect this global nature.

\subsubsection{Interpretation-Driven Training.}\label{subsubsec:interpretation driven training}
%In this subsection we consider \emph{interpretation driven training}, an example application of explanations. %\DA{It's actually not an application right? I don't see where you compute and use the explanation map, except in the end to compare the two models} \ron{During training you compare the output to the explanation that deletes all of the buildings and keeps all of the samples. This is an explanation. Here, the mask consists of two values, similarly to the phase vs magnitude explanation.} For simplicity, we consider the inpainting of all unchosen measurements with $\Phi_{\text{gt}}$.
% %
We now discuss an example application of the explanation obtained by the RDE approach described above, called \emph{interpretation driven training}. When a missing building is in line-of-sight of a Tx, we would like $\Phi$ to reconstruct this building relying on samples in the shadow of the building rather than patterns in the  city. To reduce the reliance of $\Phi$ on the city information in this situation, one can add a regularization term in the training loss which promotes explanations relying on measurements. 
%To detect a missing building in a perturbed input city map, we can compare the output of $\Phi$ and $\Phi_{\text{gt}}$. Furthermore, if the missing building is in line-of-sight of the Tx, then the minimal value of the $\Phi_{\text{gt}}$ simulation inside the building must be higher than some fixed value. 
%\DA{Ron, am I correct about detecting a missing building?}
%\ron{Not exactly. During supervised learning, we know which buildings are missing. We just need to estimate if they are in line of sight of the Tx. Note that the network that we train does not get the missing building od the line-of-sight information as input. This input is only used to define the loss during learning. You determine if the building is missing (durin training) if $\Phi_{gt}$ is higher then some value there. However, perhaps we don't need to explain how we check line-of-sight in this application. I used this approach since it was simple for me to implement, but you can also use a geometric method that explicitly check if the building is in line-of-sight.  We just need to say that during training we know which buildings are missing, and we check if the are in line-of-sight. If so, we add a term to the loss}
\begin{figure}[h]
	\centering
	\begin{subfigure}{0.23\textwidth}\centering
		\includegraphics[width=0.997\linewidth]{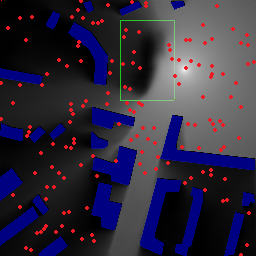}
		\caption{Vanilla $\Phi$ estimation \newline}
	   \label{fig:radio_driven1}
	\end{subfigure}%
	\hfill
	\begin{subfigure}{0.23\textwidth}\centering
		\includegraphics[width=0.997\linewidth]{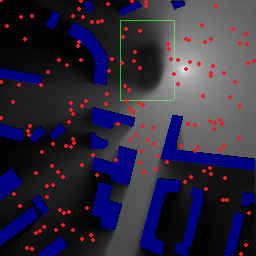}
		\caption{Interpretation-driven $\Phi_{\text{int}}$ estimation}
	   \label{fig:radio_driven2}
	\end{subfigure}%
	\hfill
	\begin{subfigure}{0.23\textwidth}\centering
		\includegraphics[width=0.997\linewidth]{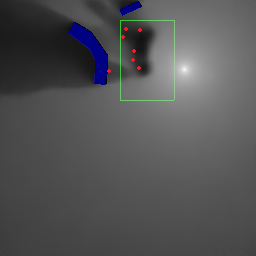}
		\caption{Vanilla $\Phi$ explanation \newline}
	   \label{fig:radio_driven3}
	\end{subfigure}%
	\hfill
	\begin{subfigure}{0.23\textwidth}\centering
		\includegraphics[width=0.997\linewidth]{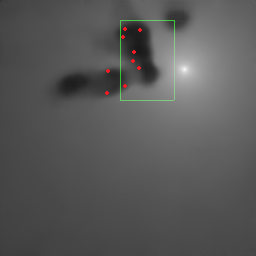}
		\caption{Interpretation-driven $\Phi_{\text{int}}$ explanation}
			\label{fig:radio_driven4}
	\end{subfigure}%
	\caption{{Radio map estimations, interpretation driven training vs vanilla training}: The radio map (gray), input buildings (blue), input measurements (red), and domain of the missing building (green box).}
	\label{fig:radio_2}
\end{figure}
Suppose $x = [x^{(0)}, x^{(1)}, x^{(2)}]$ contains a missing input building in line-of-sight of the Tx location and denote the subset of pixels of the missing building in the city map as $J_x$. Denote the prediction by $\Phi$ restricted to the subset $J_x$ as $\Phi_{J_x}$.
Moreover, define  $\tilde{x} \coloneqq [x^{(0)}, 0, x^{(2)}]$ to be the modification of $x$ with all input buildings masked out. We then define the \emph{interpretation loss} for $x$ as 
\begin{align*}
    \ell_{\text{int}}(\Phi, x) \coloneqq \norm{\Phi_{J_x}(x) - \Phi_{J_x}(\tilde{x})}_2^2.
\end{align*}
The interpretation driven training objective then regularizes $\Phi$ during training by adding the interpretation loss for all inputs $x$ that contain a missing input building in line-of-sight of the Tx location. An example comparison between explanations of the vanilla RadioUNet $\Phi$ and the interpretation driven network $\Phi_{\text{int}}$ is given in \Cref{fig:radio_2}.

\section{Conclusion}
%In this chapter we presented the \emph{Rate-Distortion Explanation} (RDE) framework to explain decisions made by black-box models such as neural networks. The framework is motivated by the rate-distortion-theory, which studies lossy data compression. In this framework, the explanation is determined as an optimal mask on the features of the input signal in the sense that the distortion between the original output and the output at the resulting obfuscated signal is minimized. As a generic framework, we have huge flexibility in choosing the way to represent signals, in generating obfuscations and in selecting a measure of the distortion. This in turn yields the flexibility of the framework to adapt to different interpretation queries and underlying tasks, as well as diverse data modalities.

In this work, we presented the \emph{Rate-Distortion Explanation} (RDE) framework in a revised and comprehensive manner. Our framework is flexible enough to answer various interpretation queries by considering suitable data representations tailored to the underlying domain and query. We demonstrate the latter and the overall efficacy of the RDE framework on an image classification task,  on an audio signal classification task, and on a radio map estimation task, a seldomly explored regression task.


\bibliographystyle{plain}
\bibliography{references}
\end{document}